 \documentclass[journal]{IEEEtran}

 \usepackage[left=0.625in,right=0.625in,top=0.75in,bottom=0.99in]{geometry}
\usepackage{cite}
\usepackage{amsmath,amssymb,amsfonts}
\usepackage{graphicx}
\usepackage{color}
\usepackage{subfig}
\usepackage{stackengine}
\usepackage{comment}
\usepackage[normalem]{ulem}
\usepackage{tabularx}
\usepackage{bm}
\usepackage{dsfont}
\usepackage{mathtools}
\usepackage[font=scriptsize]{caption}

\usepackage{booktabs,multirow,array,caption}

\usepackage[singlespacing]{setspace} 
\usepackage[colorlinks, citecolor=blue, bookmarks=false]{hyperref} 

\usepackage[flushleft]{threeparttable} 
\usepackage{booktabs}

\usepackage{caption}

\usepackage{textcomp}
\usepackage{xcolor}

\usepackage[algo2e]{algorithm2e} 
\SetKwRepeat{Do}{do}{while}%

\SetCommentSty{mycommfont}
\usepackage{algorithm}
\usepackage{algpseudocode}

\def\BibTeX{{\rm B\kern-.05em{\sc i\kern-.025em b}\kern-.08em
    T\kern-.1667em\lower.7ex\hbox{E}\kern-.125emX}}
    
\newcommand{\noteminor}[1]{\textcolor{black}{{ #1}}}

\algnewcommand\algorithmicforeach{\textbf{for each}}
\algdef{S}[FOR]{ForEach}[1]{\algorithmicforeach\ #1\ \algorithmicdo}

\usepackage{orcidlink}

\usepackage{xspace}
\usepackage{ifthen}
\usepackage[inline]{enumitem}

\usepackage[utf8]{inputenc}

\usepackage{algcompatible}
\usepackage{algorithmicx}

\newboolean{showcomments}
\setboolean{showcomments}{true}
\ifthenelse{\boolean{showcomments}}
{
}

\usepackage{subfig}
\usepackage{fancyhdr}
\usepackage{amssymb}
\usepackage{pifont}

\newcommand{\cmark}{\textcolor{green}{\ding{51}}} 
\newcommand{\xmark}{\textcolor{red}{\ding{55}}}   

\begin{document}

\renewcommand{\algorithmiccomment}[1]{\textit{#1}}

\title{VISTA: A Framework for Vision-Informed Semantic Traffic Awareness}
\author{
\IEEEauthorblockN{Murat Arda Onsu$^1$, Poonam Lohan$^1$, Burak Kantarci$^1$, Aisha Syed$^2$, Matthew Andrews$^2$, Sean Kennedy$^2$}\\
\IEEEauthorblockA{\textit{$^1$University of Ottawa, Ottawa, ON, Canada}\\
\textit{$^2$Nokia Bell Labs, 600 March Road,
Kanata, ON K2K 2E6, Canada}\\
$^1$\{monsu022, ppoonam, burak.kantarci\}@uottawa.ca,~$^2$\{aisha.syed, matthew.andrews, sean.kennedy\}@nokia-bell-labs.com}
}

\maketitle

\begin{abstract}
The Internet of Things (IoT) is transforming intelligent transportation systems by connecting roadside cameras, edge computing units, vehicles, and traffic infrastructure into a shared sensing and decision-making ecosystem. However, transmitting raw traffic video over vehicle-to-everything (V2X) links is bandwidth-intensive, while deploying large multimodal models directly on resource-constrained vehicles is often impractical. To address these limitations, this article proposes VISTA (Vision-Informed Semantic Traffic Awareness), a multimodal vision-language edge-intelligence framework that combines predictive video representation learning with lightweight language-based reasoning to provide communication-efficient traffic situational awareness. In the proposed architecture, roadside surveillance units employ the Video Joint Embedding Predictive Architecture (V-JEPA) and an attentive-pooler module to convert traffic video clips into compact future-scene embeddings. These embeddings are transmitted through V2X links to vehicle-mounted systems, where a TinyLlama small language model, enhanced with a custom visual adapter, interprets the embeddings alongside natural-language (NL) traffic queries.  VISTA enables vehicles to obtain interpretable traffic information, reason about upcoming scene dynamics, and identify potential collision risks without receiving raw video streams. Experimental results in a digital-twin urban traffic environment show that VISTA achieves 97.77\% accuracy and an 89\% F1 score for future collision prediction while reducing the communication payload by several orders of magnitude compared with raw RGB video transmission. These results demonstrate the feasibility of embedding-based multimodal intelligence for low-latency, communication-efficient, and safety-oriented IoT-enabled transportation systems.
\end{abstract}


\section{Introduction} \label{sec:1}
The rapid deployment of Internet of Things (IoT) technologies is transforming intelligent transportation systems (ITS) into connected sensing and decision-making environments. Roadside cameras, traffic signals, roadside units (RSUs), connected vehicles, and edge devices continuously generate heterogeneous data that support traffic monitoring, congestion mitigation, and reduced collision likelihood \cite{dt}. In IoT-enabled urban mobility systems, traffic intelligence must go beyond passive observation by interpreting scenes, anticipating events, and providing timely information to vehicles and traffic operators. Traditional approaches rely on classical machine learning, including vehicle detection and heuristic rules. While useful in controlled settings, these methods often struggle in dynamic urban environments including changing lighting, weather, and driver behaviors. An effective traffic-safety system should detect road users, understand their interactions, answer traffic queries, and identify collision risks proactively. 


Recent language model (LM) advances provide interpretable and adaptive traffic intelligence. LLMs' ability to reason in context and communicate in NL enables adaptation and interpretation of intelligence in urban traffic control systems \cite{4}.  Large LMs (LLMs) require significantly more memory, computation, and energy, making their deployment on resource-constrained RSUs and in-vehicle edge devices difficult. TinyLlama, Phi-3, and OpenELM are practical SLMs that reduce computational requirements while maintaining reasoning capabilities \cite{5}. Most SLMs are text-only models and cannot process visual traffic data, limiting their use in safety-critical ITS applications. Integrating visual perception and language-based reasoning requires multimodal integration \cite{rel,15}. Allowing multimodal intelligence in IoT-enabled ITS presents a major communication challenge. Transmission of raw video streams from RSUs or surveillance cameras to vehicles over V2X links can increase bandwidth and latency, especially in dense urban environments with many connected vehicles and cameras. Practical traffic intelligence requires accurate perception, reasoning, and efficient information exchange.


To address these challenges, we propose VISTA (Vision-Informed Semantic Traffic Awareness), a multimodal edge-intelligence framework that enables predictive scene understanding and language-guided traffic reasoning through compact semantic video representations. Unlike our previous ViT-based approach \cite{onsu2025semantic}, which generates image-based embeddings for current-scene understanding, VISTA uses V-JEPA with future-oriented training to capture spatiotemporal dynamics and generate compact embeddings for future-scene reasoning and collision prediction. Predicting future scene likelihoods using these embeddings supports intelligent traffic tasks such as collision anticipation and proactive road safety planning \cite{9}. These embeddings are processed by an attentive-pooler module to generate a single future-scene representation that captures task-relevant near-future traffic information. Instead of transmitting raw video, the RSU shares this compact embedding through V2X links with nearby vehicles. On the vehicle side, a TinyLlama SLM \cite{11} equipped with a custom visual adapter processes the received embedding together with NL traffic queries to provide interpretable responses and future collision-risk predictions.
The main contributions of this work under the VISTA framework are summarized as follows:
\begin{itemize}
\item A V-JEPA-based edge video encoding module is employed at RSUs to generate compact traffic embeddings and reduce raw video transmission over V2X links.
\item An attentive-pooler-based future embedding module is developed to compress V-JEPA patch-level outputs into a single future-scene representation for proactive traffic-safety analysis.
\item A multimodal RSU-to-vehicle intelligence pipeline is introduced, where future-oriented embeddings generated from 16-frame clips (sampled at a 4-fps rate over a 4-second temporal window) are interpreted by the TinyLlama SLM with a custom visual adapter for NL traffic queries and collision-risk prediction, with the future prediction target defined as the last frame of the next second.
\end{itemize}
The proposed framework, VISTA, achieves 97.7\%  accuracy and 89\% F1 score for collision prediction, and demonstrates strong prediction performance and reduced communication payload through compact embedding transmission.


\section{Use Cases and Requirements of Intelligent Traffic Monitoring} \label{sec:2}

Intelligent traffic monitoring (ITM) is vital for IoT-enabled ITS, where roadside cameras, RSUs, and edge devices supporting real-time traffic management and safety. Unlike traditional systems that mainly record visuals or trigger rule-based alerts, modern ITM systems extract semantic knowledge, predict near-future risks, and provide actionable information to drivers, vehicles, and operators.


\renewcommand{\arraystretch}{1.2}
\begin{table}[t]

\fontsize{9.0}{9}\selectfont
\caption{\small{Comparison of Recent ITS Research on Multimodal Traffic Monitoring (\cmark: Considered \xmark: Not considered)}}

\begin{tabular}{lllllll}
\hline
                     & \cite{15}       & \cite{7}        & \cite{12}       & \cite{new_1}    & \cite{onsu2025semantic} & {\color[HTML]{FE0000} \textbf{VISTA}}   \\ \hline
Language model       & \textit{\cmark} & \textit{\cmark} & \textit{\cmark} & \xmark          & \textit{\cmark}         & {\color[HTML]{FE0000} \textit{\cmark}} \\ \hline
Image                & \textit{\cmark} & \textit{\cmark} & \textit{\cmark} & \textit{\cmark} & \textit{\cmark}         & {\color[HTML]{FE0000} \textit{\cmark}} \\ \hline
Video                & \xmark          & \xmark          & \textit{\cmark} & \textit{\cmark} & \xmark                  & {\color[HTML]{FE0000} \textit{\cmark}} \\ \hline
Future Predictions   & \xmark          & \textit{\cmark} & \xmark          & \textit{\cmark} & \xmark                  & {\color[HTML]{FE0000} \textit{\cmark}} \\ \hline
Embeddings Tx          & \textit{\cmark} & \xmark          & \xmark          & \xmark          & \textit{\cmark}         & {\color[HTML]{FE0000} \textit{\cmark}} \\ \hline
Collision Prediction & \textit{\cmark} & \xmark          & \textit{\cmark} & \textit{\cmark} & \textit{\cmark}         & {\color[HTML]{FE0000} \textit{\cmark}} \\ \hline
Monitoring           & \textit{\cmark} & \textit{\cmark} & \textit{\cmark} & \textit{\cmark} & \textit{\cmark}         & {\color[HTML]{FE0000} \textit{\cmark}} \\ \hline
\end{tabular}
\label{my table}
\end{table}

\begin{figure*} [t]
    \centering
    \includegraphics[width=1\linewidth]{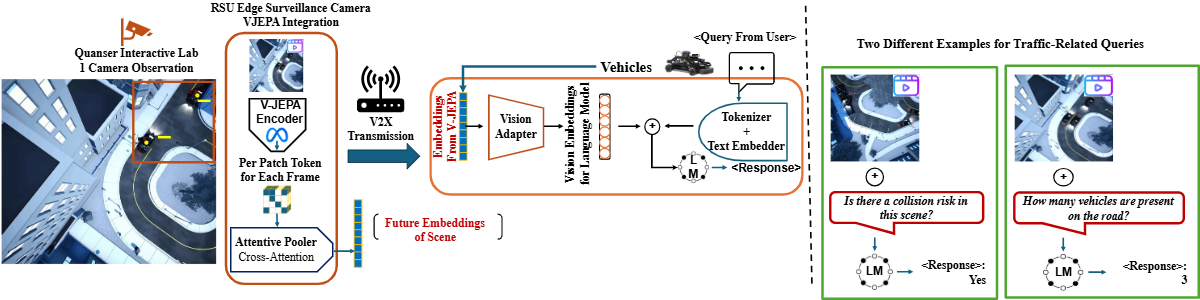}
    \caption{Overall workflow of the proposed VISTA framework}
    \label{fig:img2}
\end{figure*}

\subsection{Use Cases in Intelligent Traffic Monitoring}
The main ITM use cases fall into two categories: traffic-flow management and safety-critical event response. For traffic-flow management, continuous scene understanding enables traffic-density estimation, congestion detection, route guidance, and adaptive signal control. These functions reduce waiting time, improve road utilization, and support more efficient urban mobility.

ITM systems detect hazards like sudden braking, unsafe lane changes, pedestrian conflicts, jaywalking, or approaching vehicles at intersections. By interpreting current scenes and predicting near-future events, they can warn connected vehicles or operators before incidents happen. This is crucial in occluded urban areas where visibility is limited. \cite{8}.


\subsection{Requirements for Intelligent Traffic Monitoring}


An effective ITM system must meet several requirements to support these use cases. It needs low-latency inference for timely safety advisories and collision warnings, generated before traffic changes. Communication efficiency is crucial, as transmitting raw video over V2X and edge networks can cause overload. Goal-oriented and semantic communication can cut this by sending only task-relevant information  \cite{goal}. Scalability is essential to manage increasing data and expand coverage across intersections \cite{8}. Additionally, resource efficiency, multimodal understanding, adaptability, reliability, and privacy are vital. AI models on RSUs, cameras, or in-vehicle processors must work within limited resources, integrating visual data, vehicle states, infrastructure info, maps, and user queries for clear traffic insights. The system must also adapt to weather, lighting, road layouts, and driver behavior, preventing delays, errors, and unnecessary data sharing.

 
Table~\ref{my table} compares recent ITS studies according to their support for LMs, image/video inputs, future prediction, embedding transmission, collision prediction, and monitoring. Existing works have investigated multimodal LMs for image-based traffic understanding and collision prediction \cite{15,onsu2025semantic}, future-state prediction \cite{7}, video-based traffic analysis \cite{12,new_1}, and embedding-based communication \cite{15,onsu2025semantic}. However, few works jointly consider predictive video embeddings, communication-efficient V2X transmission, and language-model-based traffic reasoning. This motivates the proposed framework, VISTA, which aims to provide low-latency and interpretable traffic monitoring by transmitting compact future-scene embeddings instead of raw video streams.

\section{Proposed Framework: Vision-Informed Semantic
Traffic Awareness (VISTA)} \label{sec:3}

VISTA consists of two main components: (i) RSU-level predictive video embedding generation and (ii) vehicle-level multimodal reasoning. At the RSU, traffic videos captured by edge surveillance cameras are encoded into compact future-scene embeddings using V-JEPA and an attentive-pooler module. These embeddings are then transmitted through V2X links to nearby vehicles. At the vehicle, a TinyLlama-based SLM, enhanced with a custom visual adapter, combines the received embedding with textual traffic queries to generate interpretable responses and collision-risk predictions. Fig.~\ref{fig:img2} shows the overall workflow.

\subsection{Dataset Collection and Digital Twin Environment} \label{data}

Traffic video records are collected using the Quanser Interactive Lab (Qlab)\footnote{Quanser, *Quanser Python API*, version 2024.10.17, 2024. Available at: \url{https://docs.quanser.com/quarc/documentation/python/getting_started.html}}, a configurable digital-twin environment for safe and repeatable ITS evaluation \cite{dt}. It includes vehicles, pedestrians, traffic lights, signs, roads, sidewalks, and surveillance cameras controlled through Python or MATLAB. Three camera views capture diverse traffic scenarios, including intersections and four-way junctions. The dataset\footnote{Urban Mobility Research Dataset, Video Records (Generated with the Quanser Interactive Lab), Available at: \url{https://dx.doi.org/10.21227/jazm-2z18}} contains 1975 urban traffic video records, each lasting 4 to 8 seconds. Original videos are captured at $2048 \times 2048$ resolution and 20 fps. Then, each video record is processed by cropping the most relevant road region and resizing it to $256 \times 256$. 
This process is followed by reducing the video fps rate to
minimize the number of input frames. The 4 fps sampling rate is selected to match the lightweight V-JEPA input configuration and reduce inference time. Higher sampling rates can capture finer temporal changes but increase the number of processed frames, memory usage, computational cost, and latency.
For future-frame prediction, the target future frame is defined as the last frame of the next second, allowing the model to learn near-future traffic dynamics while maintaining a short prediction horizon suitable for safety-critical traffic monitoring.

\subsection{Video Joint Embedding Generation} \label{vjepa}
\begin{figure*} [t]
    \centering
    \includegraphics[width=0.99\linewidth]{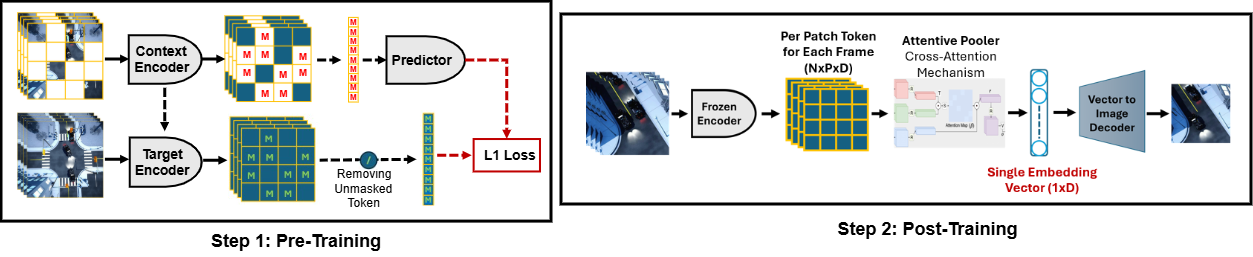}
    \caption{V-JEPA module of the proposed VISTA framework: pretraining (Step 1) and future-scene embedding generation (Step 2).}
    \label{fig:img4}
\end{figure*}


V-JEPA is employed at the RSU to extract predictive spatiotemporal representations from traffic videos \cite{14}. Video models like TimeSformer and VideoMAE excel at understanding visual content but focus on current frames. V-JEPA, based on predictive learning, estimates masked or future latent regions from context, making it better suited for anticipatory tasks like future-scene understanding and collision prediction. It is pretrained via self-supervised masked prediction.. Training steps, which are illustrated on the left part of \figurename \ref{fig:img4}, include three networks: context encoder, target encoder, and predictor. Each frame of the video clip, which has spatial dimensions $H \times W$, is divided into non-overlapping spatiotemporal, $P=\tfrac{H}{p} \times \tfrac{W}{p}$, patches, determined by a patch size of $p \times p$. The target encoder processes the full unmasked clip, while the context encoder encodes a masked variant. A predictor network generates embeddings for the masked tokens, where each patch is mapped into a $D$-dimensional embedding vector, yielding a per-frame embedding tensor of dimensions $P \times D$. Then these embeddings are trained against target encoder embeddings using an $L_1$ loss. This allows the model to learn predictive spatiotemporal representations in the embedding space. 
However, for $N$ frames per clip, generating embeddings tensors of dimension $N \times P \times D$  still consumes more memory, and it must be converted into single embeddings tensor of dimension $1 \times D$ that map the most essential information and future predictions before transmission.

\subsection{Future-Frame Embedding Prediction} \label{embeddings}

Pretrained V-JEPA model, with frozen encoder, is employed from the Section \ref{vjepa}, which receives video records and generates a per-patch token for each of its frames $N \times  P \times D$. In the training for future embedding generation step, which is illustrated on the right part of \figurename \ref{fig:img4}, an attentive pooler network is deployed alongside the Vector-to-Image (V2I) Decoder. The attentive pooler uses learnable query tokens and cross-attention to aggregate the V-JEPA embeddings into a single future-scene vector. Let $\mathbf{Z} \in \mathbb{R}^{(N \cdot P) \times D}$ denote the V-JEPA token embeddings and $\mathbf{Q} \in \mathbb{R}^{q \times D}$ denote the learnable query matrix, ($q=1$ in this work). The pooled representation is computed as
\begin{equation}\small
\mathbf{z}_{\mathrm{attn}} =
\mathrm{softmax}\left({\mathbf{Q}\mathbf{Z}^{\top}}/{\sqrt{D}}\right)\mathbf{Z}.
\end{equation}

During training, the V2I decoder reconstructs the next sampled frame from $\mathbf{z}_{\mathrm{attn}}$, encouraging the compact embedding to preserve near-future traffic information. The training objective combines L1, SSIM, and LPIPS losses. Since the background is mostly fixed, the L1 loss is computed mainly over vehicle and pedestrian regions detected by YOLOv11 to emphasize dynamic road users rather than static scenery. To balance pixel-level, structural, and perceptual reconstruction quality, Adaptive Loss Training is applied where the three loss terms are initially assigned equal weights, and their weights are periodically adjusted according to their relative validation losses. After training, the V2I decoder is discarded, and the attentive-pooler output is used as the transmitted future-scene embedding.

\subsection{SLM Deployment at Vehicle for Traffic-Related Queries} \label{llama}

TinyLlama \cite{11} operates as the SLM backbone because of its 1.1B parameters and 2.2 GB memory footprint, making it suitable for low-resource vehicle units. It processes NL queries and future-scene embeddings from the RSU to generate traffic responses. Since TinyLlama is text-only, a custom visual adapter projects the embeddings into visual tokens, normalizes them, and processes them through an MLP, producing tokens aligned with the SLM input space. These tokens are concatenated with textual query embeddings and fed to TinyLlama. During training, TinyLlama stays frozen, and only the adapter updates, reducing training costs and enabling collision-risk prediction and traffic interpretation.


NL queries in VISTA are not intended only for human-interaction, they also provide a flexible task-specification interface for ITM. Instead of designing a separate formal interface for each traffic objective, system designers can describe the required information directly in NL while reusing the same transmitted future-scene embedding. This allows the framework to support different operational needs without modifying the embedding-generation and communication pipeline. At deployment time, the SLM output can still be constrained to structured responses, such as binary risk warnings, numerical counts, or categorical advisories, so automated vehicle-control, route-planning, or traffic-control modules can consume the results. Extending VISTA to new query types like blocked-lane detection, pedestrian presence, congestion, unsafe maneuvers, or intersection clearance requires task-specific annotations, training, evaluation, safety validation, and is left for future work.

\subsection{Computational and Communication Trade-Off Analysis}

IoT-enabled ITS must balance predictive performance with the computational and communication constraints of RSU and in-vehicle devices. In a conventional architecture, raw video streams captured by RSU surveillance cameras would need to be transmitted directly to vehicles or centralized servers for further processing. For a raw RGB video clip with $N$ frames, resolution $H \times W$, $C$ channels, and $b$ bytes per pixel, the payload is
\begin{equation}
S_{\text{video}} = N \times H \times W \times C \times b .
\end{equation}
Transmitting raw video therefore introduces significant bandwidth consumption and transmission latency, especially in dense traffic environments where multiple cameras and vehicles operate simultaneously. In contrast, the proposed pipeline transmits a compact embedding of dimension $D$, whose payload is $S_{\text{emb}} = D \times b_e$,  where \(b_e\) denotes the number of bytes used to represent each embedding element. Since \(S_{\text{emb}} \ll S_{\text{video}}\), the proposed embedding-based transmission significantly reduces communication overhead.

From a computational perspective, this design introduces additional processing at the edge infrastructure. Let \(C_{\text{enc}}\) denote the computational cost of the V-JEPA encoder, which scales with the number of frames and patches, typically on the order of $\mathcal{O}(N \times P \times D)$, while the attentive pooler introduces an additional cross-attention operation whose complexity depends on the number of query  tokens \(q\),
$\mathcal{O}(q \times N \times P \times D)$. Though this increases RSU computation, it removes the need for vehicles to run costly vision models locally. Vehicles instead process only compact embedding vectors and textual queries with TinyLlama. This shifts the computational load from vehicles to shared edge infrastructure and greatly reduces communication requirements.


\subsection{RSU-to-Vehicle Communication and E2E Latency Analysis}


Each RSU links to a local traffic region, like an intersection or road segment with edge cameras. It periodically broadcasts the future-scene embedding to vehicles within its communication region. Each message includes the RSU identifier, scene or camera identifier, timestamp, coverage region data, validity horizon, and embedding vector. These metadata enable vehicles to verify whether the received embedding matches their location, heading, and route, discarding outdated information. The architecture also supports a request mode, where approaching vehicles query the RSU for the latest embedding. The communication interface is compatible with existing and emerging V2X families, including DSRC/IEEE 802.11p/ITS-G5 and cellular V2X evolution from LTE/NR C-V2X toward 5G-Advanced and 6G-oriented studies in 3GPP Release~19/20 \cite{3gppr19}. This work focuses on the application-layer payload and end-to-end AI pipeline; protocol-level scheduling, congestion control, and reliability optimization are left for future system-level deployment.

For safety-critical operation, the end-to-end response time includes both AI inference and communication delay, \noteminor{which represents the sum of 
 the video acquisition/preprocessing time, RSU-side delays ($T_{\mathrm{cap}}$ and $T_{\mathrm{V\text{-}JEPA}}$), RSU-side inference delays ($T_{\mathrm{pool}}$),V2X transmission delay ($T_{\mathrm{tx}}$), and the vehicle-side processing delays ($T_{\mathrm{adapter}}$ and $T_{\mathrm{SLM}}$)}. On the other hand, the communication latency Tx is calculated under different conditions such as rate congestion and packet loss \noteminor{as the ratio of the communication payload ($L$) to the product of the link rate ($R$), one's complement of the congestion level ($\rho$) and one's complement of the packet loss probability ($p$)}. 
 This formulation clarifies that real-time feasibility depends on the complete edge-to-vehicle pipeline, not only on model inference. The SLM output in vehicle functions as a safety advisory, mapping to a human-driver warning interface or used by an autonomous vehicle’s decision-making system.


\section{Performance Evaluation} \label{sec:4}

\subsection{Evaluation Setup and Parameters}

As mentioned in Section \ref{data}, with three different cameras of Qlab, 1975 videos are recorded as a dataset.
The dataset is split into 1571 training clips and 404 test clips, i.e., approximately 80\%:20\% split. Samples from all three camera views are included in both sets, while predefined vehicle trajectories and duplication checks are used to reduce overlap between highly similar samples.
Collision events are labeled using the QLabs collision-detection function. The model receives a 4-second input clip ending approximately 1 second before the collision, so the collision itself is not included in the input frames. V-JEPA uses these preceding frames to generate a predictive future-scene embedding for the upcoming traffic state. 
One of the lightest V-JEPA variants is selected, featuring a Vision Transformer with a large architecture and 1024-dimensional embedding vectors. Video clips are resized to a resolution of 256 and trimmed to 16 frames per clip. The V2I decoder consists of 2 Dense and 3 convolutional layers. It takes the attentive pooler embeddings and converts them into RGB images of the future frame. A Deep Learning model for calculating LPIPS loss is selected as the VGG model. For TinyLlama, the \textit{TinyLlama-1.1B-Chat-v1.0} version is selected, which has a 2048-dimensional input embedding. The adapter uses Transformer Encoder Layers and bridges V-JEPA embeddings to the TinyLlama embedding format. During the TinyLlama adapter training, the AdamW optimizer is used for 20 epochs. 
Models are fine-tuned on a dual-socket Intel Xeon Gold 5218R CPU system with 80 logical CPUs, 40 physical cores, and a maximum CPU frequency of 4.0 GHz, while experiments are conducted on 13th Gen Intel(R) Core(TM) i9-13900HX with 24 cores, 32 logical CPUs, and a CPU frequency of 2.2 GHz.



For the transmission-size analysis, we compare the payload required to transmit a raw RGB video clip with that of the proposed compact future-scene embedding. A 4-second RGB clip with a resolution of $256 \times 256$ at 4 fps contains 16 frames, resulting in a raw payload of $16 \times 256 \times 256 \times 3$ uint8 pixels, equivalent to approximately 3.15 MB. In contrast, the final 1024-dimensional embedding requires only 4 KB when represented in float32 and 2 KB when represented in float16. To evaluate the embedding compression result with baseline compression methods, MPEG-2, MPEG-4, H.264/AVC, and H.265/HEVC are used for comparison. According to the results, the communication payload is reduced by 13.25X, 10.82X, 6.23X, and 4.14X when embedding is presented as float32, and 26.50X, 21.63X, 12.45X, and 8.28X when it is float16.


To further evaluate real-time feasibility, we profile the complete RSU-to-vehicle processing pipeline rather than considering model inference alone. These latency values are measured under the hardware setup described above. The video acquisition and preprocessing stage requires $T_{\mathrm{cap}}$=16 ms, while V-JEPA encoding at the RSU introduces $T_{\mathrm{V\text{-}JEPA}}$=105.78 ms. The attentive-pooler module adds only $T_{\mathrm{pool}}$=2.26 ms to compress the V-JEPA patch-level tokens into a single future-scene embedding. 
The float32 embedding is evaluated at V2X rates of 1-50 Mbps, congestion levels of 0-75\%, and packet-loss rates of 0-10\%, with 50 transmission samples averaged per condition. The resulting transmission delay ranges from 0.655 ms to 145.636 ms across the considered operating conditions. Assuming an effective V2X data rate of 10 Mbps, the corresponding transmission delay is $(4096 \times 8)/(10 \times 10^{6})$, or approximately $T_{\mathrm{tx}}$=3.27 ms, and transmitting the 3.15 MB uncompressed RGB video requires approximately 2.5 s. In comparison, the proposed approach requires \(T_{\mathrm{cap}}+T_{\mathrm{V\text{-}JEPA}}+T_{\mathrm{pool}}+T_{\mathrm{tx}}=127.31\) ms for RSU-side processing and embedding transmission, which represents approximately a 94.9\% reduction relative to the raw-video transmission time.

On the vehicle side, the visual adapter requires $T_{\mathrm{adapter}}$=4.55 ms to map the received embedding into the TinyLlama-compatible token space, while TinyLlama response generation requires $T_{\mathrm{SLM}}$=73 ms. Under the experimental hardware and communication assumptions, the edge-to-vehicle pipeline has an end-to-end latency of about 204.86 ms. This shows that VISTA can deliver near-real-time traffic safety advisories without the bandwidth use of raw video transmission.

\begin{table}[]
\centering 
\caption{\small{Training Setup and Parameters}}
\label{tab:parameters}
\fontsize{8.0}{10}\selectfont
\begin{tabular}{|l|l|}\hline
Training Video   Clips & 1571                       \\ \hline
Testing Video Clips    & 404                        \\ \hline
Frame Per Clip         & 16                         \\ \hline
Video Size (WxH)       & 256                        \\ \hline
V-JEPA                 & vjepa2-vitl-fpc16-256-ssv2 \\ \hline
TinyLlama              & TinyLlama-1.1B-Chat-v1.0   \\ \hline
D$_{VJEPA}$ output     & 1024                       \\ \hline
D$_{V2I}$ input        & 1024                       \\ \hline
D$_{TinyLlama}$ Input  & 2048                       \\ \hline
Model for LPIPS-loss   & VGG                        \\ \hline

Optimizers             &    AdamW                   \\ \hline
Learning rate          & 5e-5                       \\ \hline

Epoch: V2I, Attentive Pooler       & 800              \\ \hline
Epoch: TinyLlama         & 20                         \\ \hline

\end{tabular}
\end{table}

\begin{figure} []
    \centering
    \includegraphics[width=0.99\linewidth]{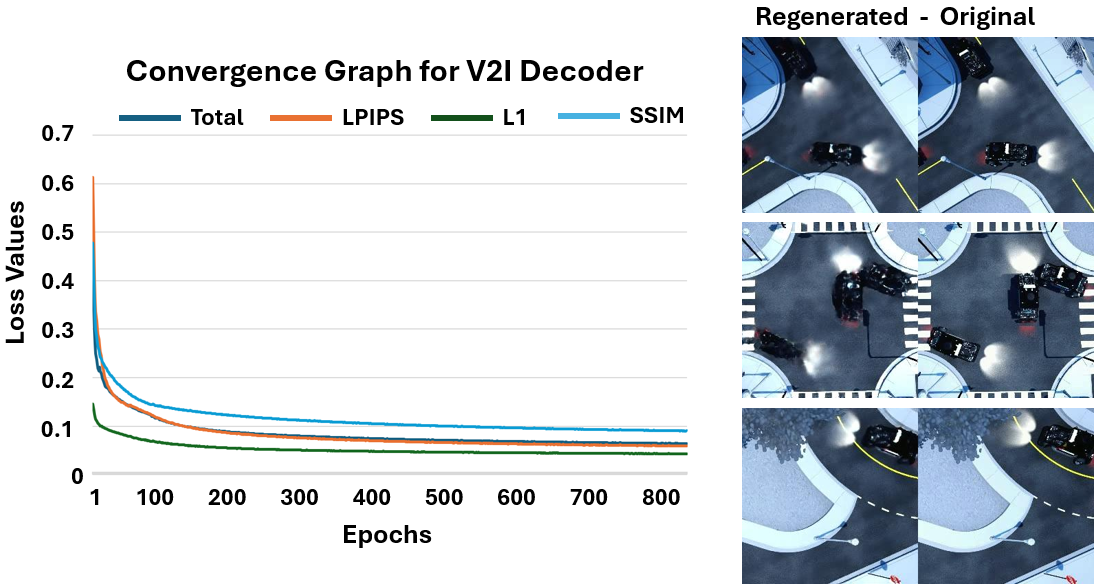}
    \caption{LPIPS, L1, SSIM and total Losses for Next Frame Generation (On Left) and Comparison of Original Next Frame and Regenerated (On Right)}
    \label{fig:img5}
\end{figure}

\subsection{Future Frame Embeddings Results}


Training results for future-frame generation improve across all performance metrics. The V2I decoder trained for 800 epochs shows the total loss declining from 0.41 at epoch 1 to 0.064 at the last epoch, indicating accurate reconstructions. LPIPS and L1 losses drop from 0.61 and 0.15 to 0.059 and 0.043, reflecting better perceptual and pixel similarity. SSIM loss decreases from 0.47 to 0.09, showing reduced reconstruction error. The steady decline confirms stable training and effective learning of temporally coherent future frames. Overall, the results demonstrate VISTA's convergence for future-frame synthesis. \figurename~\ref{fig:img5} illustrates the loss trends on the left and compares the regenerated future frames with the original frames on the right.

\subsection{Language Model Performance Evaluation}

\begin{figure*} [t]
    \centering
    \includegraphics[width=0.9\linewidth]{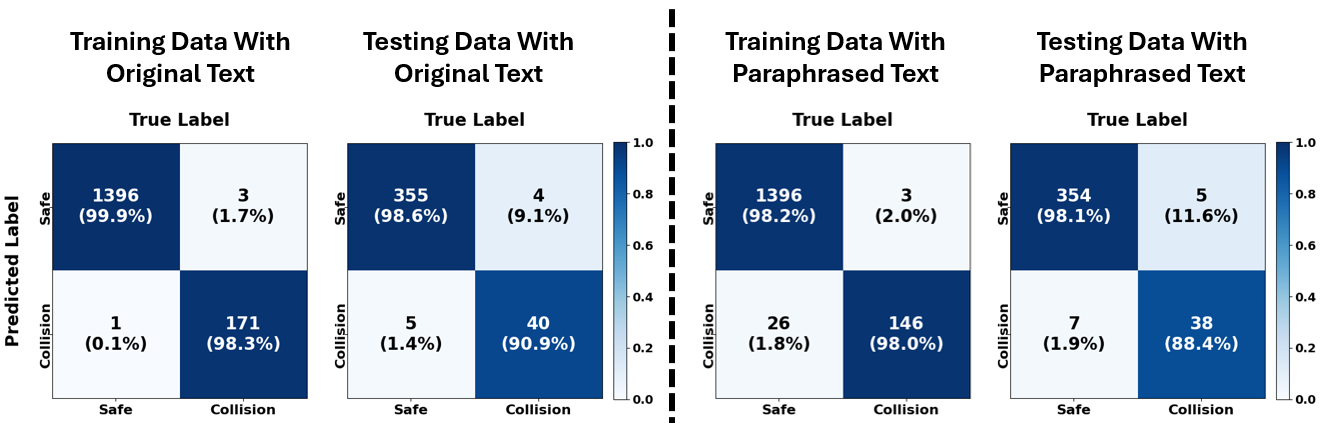}
    \caption{Collision prediction results of the proposed VISTA framework. Confusion matrices for original text queries (left) and paraphrased queries (right) illustrate the framework's robustness to variations in natural-language query formulations.}
    \label{fig:img6}
\end{figure*}

The TinyLlama-based multimodal model is tested on traffic queries: future collision-risk prediction and vehicle-count estimation. The model uses a compact future-scene embedding and a textual query to generate responses. Two evaluation settings assess performance and linguistic robustness. In the first setting, the original query templates used during training are applied. In the second setting, semantically equivalent paraphrased queries are used to test the model under more realistic variations in user wording. The corresponding confusion matrices for collision prediction are shown in Fig. \ref{fig:img6}.

The left part of \figurename \ref{fig:img6} shows the training and testing performance considering original queries, which are used in the fine-tuning, while the right half of this figure illustrates the results with paraphrased texts. For collision-risk prediction, the model performs binary classification, where collision cases are labeled as positive samples (1) and safe cases as negative samples (0). Here, the positive class corresponds a collision event involving physical contact between vehicles. A positive prediction triggers a collision warning, while a negative prediction indicates no immediate risk. For collision-risk prediction with original queries, the model achieves strong performance on both training and testing samples. In the training set, 1567 out of 1571 cases are correctly classified, with only three false negatives and one false positive. In the testing set, 395 out of 404 cases are correctly classified, with four false negatives and five false positives. This corresponds to 99.75\% training accuracy and 97.77\% testing accuracy. The model also achieves F1 scores of 99\% and 89\% for training and testing, respectively. In terms of precision and recall, the training results reach 99.42\% and 98.28\%, while the testing results achieve 88.89\% precision and 90.91\% recall. These results show that the proposed embedding-adapter-SLM pipeline can effectively interpret future-scene embeddings and identify upcoming collision risks.

The model also demonstrates promising generalization when evaluated with paraphrased queries. Although these queries use different wording and include more diverse sentence structures, the model maintains high collision-prediction accuracy. In the training set, 1542 samples are correctly classified, with three false negatives and 26 false positives, resulting in 98.15\% accuracy and a 90.97\% F1 score. In the testing set, the model correctly classifies 392 samples, with five false negatives and seven false positives, achieving 97.03\% accuracy and an 86.36\% F1 score. The small performance reduction compared with original queries indicates that the model remains robust under linguistic variation, despite being trained with limited query diversity.

For vehicle-count estimation, the model achieves 98.28\% training accuracy and 91\% testing accuracy with original queries. Under paraphrased queries, the corresponding accuracies are 89.37\% and 85.15\%. This task is more challenging than binary collision prediction because vehicle counting requires finer scene-level interpretation, especially when vehicles are partially visible, occluded, or located near image boundaries. Nevertheless, most incorrect predictions differ from the ground truth by only one vehicle, indicating that the model generally preserves useful scene awareness. Overall, the results confirm that compact future-scene embeddings can be effectively interpreted by a lightweight SLM for traffic-safety reasoning. The high collision-prediction accuracy under both original and paraphrased queries demonstrates the feasibility of using SLM-based multimodal reasoning for communication-efficient and interpretable ITS applications. The results also suggest that further gains can be achieved by expanding the training set with more diverse paraphrased traffic queries and scene descriptions.

Moreover, to demonstrate the contribution of the VISTA components, separate ablation studies are conducted for attentive pooling and the TinyLlama visual adapter against alternatives. With the adapter architecture fixed, attentive pooling improves vehicle-count accuracy by 11.82\% and 16.83\% over mean and max pooling, respectively. With attentive-pooled features fixed, the original visual adapter improved vehicle-count accuracy by 8.84\% and 4.64\% over linear and two-layer MLP adapters, respectively, while improving collision F1 by 6.08\% and 32.54\%. These results show that both attentive pooling and the proposed visual adapter substantially contribute to the overall performance, and they are effective in extracting and transforming task-relevant visual representations.

\section{Opportunities and Challenges} \label{sec:5}
LMs can enhance ITS by combining visual embeddings, sensor data, and textual queries to support interpretable traffic monitoring, safety warnings, congestion management, and adaptive control. However, practical deployment still faces challenges in computation, latency, privacy, reliability, and robustness, as discussed below:

\subsection{Cost and Resource Requirement}

Multimodal LMs introduce high computational cost, memory usage, and energy consumption, particularly when deployed continuously at the edge. Large models rely on billions of parameters, yet most RSU and in-vehicle processors lack the hardware capacity to run inference at the required speed, causing delays or accuracy degradation in resource-constrained environments \cite{4}. Continuous operation for real-time perception also increases power usage and maintenance overhead, while frequent model updates and synchronization across distributed nodes add bandwidth and storage burdens. This creates a fundamental trade-off between model complexity and deployment feasibility in large-scale ITS.
At the same time, this challenge opens several research opportunities. Model compression, parameter-efficient fine-tuning, distillation, sparse attention, quantization, and low-rank adaptation can reduce the footprint while preserving accuracy. Collaborative inference between vehicles and RSU offers another promising direction, enabling distributed processing that lowers computation overhead and supports cost-efficient deployment at scale.

\subsection{Hallucination in Language models}

Hallucination is a major concern when deploying LMs in safety-critical ITS applications. In this case, the model may generate a response that sounds reasonable but is incorrect, such as misreporting vehicle movements, traffic-light states, pedestrian actions, or collision risks \cite{5}. Such errors can occur due to model bias, limited context, quantization, domain shift, or weak alignment between visual embeddings and textual reasoning. In traffic monitoring, even a small incorrect response may lead to delayed warnings or unsafe decisions. This challenge highlights the need for stronger grounding and verification mechanisms, such as uncertainty estimation, cross-checking responses with visual evidence, rule-based safety constraints, and multimodal datasets with explainable labels and temporal consistency.

\subsection{Complete Real-Time Integration Processing}

Achieving full real-time integration of multimodal LMs in ITS remains challenging due to strict latency, synchronization, and reliability requirements. Continuous video streams and textual inputs must be processed simultaneously, yet the sequential nature of transformer models and the computational cost of multimodal fusion introduce delays that hinder fast decision-making. 
The distributed structure of ITS spanning edge cameras, RSUs, and in-vehicle units further complicates network-wide synchronization. Even small delays can produce outdated responses and reduce the reliability of safety-critical tasks such as collision prediction or signal control.
This limitation creates opportunities for advancing real-time traffic intelligence. Future research may explore edge-optimized architectures, asynchronous or event-triggered inference, streaming-aware processing, and predictive scheduling that prioritizes urgent traffic events.

\subsection{Ethical and Privacy Concern}

The use of multimodal LMs in ITS introduces important privacy, accountability, and transparency concerns \cite{rel}. Traffic cameras and connected vehicles may capture sensitive information, including faces, license plates, pedestrian behavior, travel patterns, and user queries, creating risks of misuse or re-identification. These risks can reduce public trust, especially for vulnerable users. At the same time, they create opportunities for privacy-preserving ITS design, including federated learning, on-device processing, identity masking, encrypted inference, differentially private learning, and transparent model pipelines that support both regulatory compliance and reliable LM-driven traffic intelligence.

\subsection{Performance Under Variational Textual Queries}
In safety-critical applications, even small misunderstandings may lead to delayed or incorrect responses. Our results show that paraphrased queries slightly reduce accuracy and F1 scores, indicating the need for stronger linguistic robustness. Vehicle counting is more sensitive to paraphrased queries because it requires finer scene-level interpretation than binary collision prediction. Most incorrect counts differ from the ground truth by only one vehicle, particularly in scenes with partial visibility, occlusion, or vehicles near image boundaries. This also opens opportunities for paraphrase-augmented training, domain-adaptive fine-tuning, query normalization, and stronger grounding between textual queries and visual traffic evidence to improve reliability in real-world ITS deployments.

\section{Conclusions and Future Work} \label{sec:6}

This article has presented VISTA framework to illustrate how multimodal edge intelligence can transform traffic safety by combining predictive visual representations, semantic communication, and language-guided reasoning within IoT-enabled transportation systems. A key challenge in real-time traffic monitoring is not only recognizing current events but predicting what comes next and communicating that information in an interpretable form. VISTA addresses this question by using V-JEPA at edge cameras to produce compact, predictive embeddings, which are then interpreted by TinyLlama, a lightweight multimodal language model capable of answering traffic queries. The system achieved 97.77\% accuracy and an 89\% F1 score in future collision prediction and remained effective under paraphrased queries, demonstrating that SLMs can operate reliably under edge constraints. By combining predictive vision with NL reasoning, VISTA shifts ITS from passive perception to proactive safety support. Current digital-twin evaluation demonstrates the feasibility of the proposed integration, while validation under larger-scale and more diverse real-world traffic conditions is planned as part of our future work.



\section*{Acknowledgment }\label{sec:6}
This work is supported in part by the MITACS Accelerate Program, in part by the NSERC CREATE TRAVERSAL program, and in part by the Ontario Research Fund-Research Excellence program under grant number ORF-RE012-026. The authors would like to thank Quanser for their support in the generation of the traffic data via Quanser Interactive Lab.

\bibliographystyle{IEEEtran}
\bibliography{biblio.bib}{}


\end{document}